%% file: aaai2026.tex
\documentclass[letterpaper]{article}
\usepackage{aaai2026}
\usepackage{times}
\usepackage{helvet}
\usepackage{courier}
\usepackage[hyphens]{url}
\usepackage{graphicx}
\usepackage{natbib}
\usepackage{caption}
\usepackage{algorithm}
\usepackage{algorithmic}
\usepackage{newfloat}
\usepackage{listings}
\usepackage{amsmath}
\usepackage{amssymb}
\usepackage{booktabs} 

\DeclareCaptionStyle{ruled}{labelfont=normalfont,labelsep=colon,strut=off}
\floatstyle{ruled}
\newfloat{listing}{tb}{lst}{}
\floatname{listing}{Listing}

\title{MultiMotion: Multi Subject Video Motion Transfer via Video Diffusion Transformer}
\author{
    Penghui Liu\textsuperscript{\rm 1},
    Jiangshan Wang\textsuperscript{\rm 2},
    Yutong Shen\equalcontrib\textsuperscript{\rm 3},
    Shanhui Mo\equalcontrib\textsuperscript{\rm 3},
    Chenyang Qi\textsuperscript{\rm 4}\thanks{Corresponding authors.},
    Yue Ma\textsuperscript{\rm 2}\thanks{Corresponding authors.}
}
\affiliations{
    \textsuperscript{\rm 1}College of Information Science and Technology, Beijing University of Technology, Beijing, China\\
    \textsuperscript{\rm 2}Tsinghua Shenzhen International Graduate School, Tsinghua University, Shenzhen, China\\
    \textsuperscript{\rm 3}Independent Researcher\\
    \textsuperscript{\rm 4}The Hong Kong University of Science and Technology, Hong Kong\\
    penghuiliu@emails.bjut.edu.cn
}

\usepackage{bibentry}
\usepackage{subfiles}

\begin{document}

\twocolumn[{
\renewcommand\twocolumn[1][]{#1}
\maketitle
    \captionsetup{type=figure}
    \includegraphics[width=1.0\textwidth]{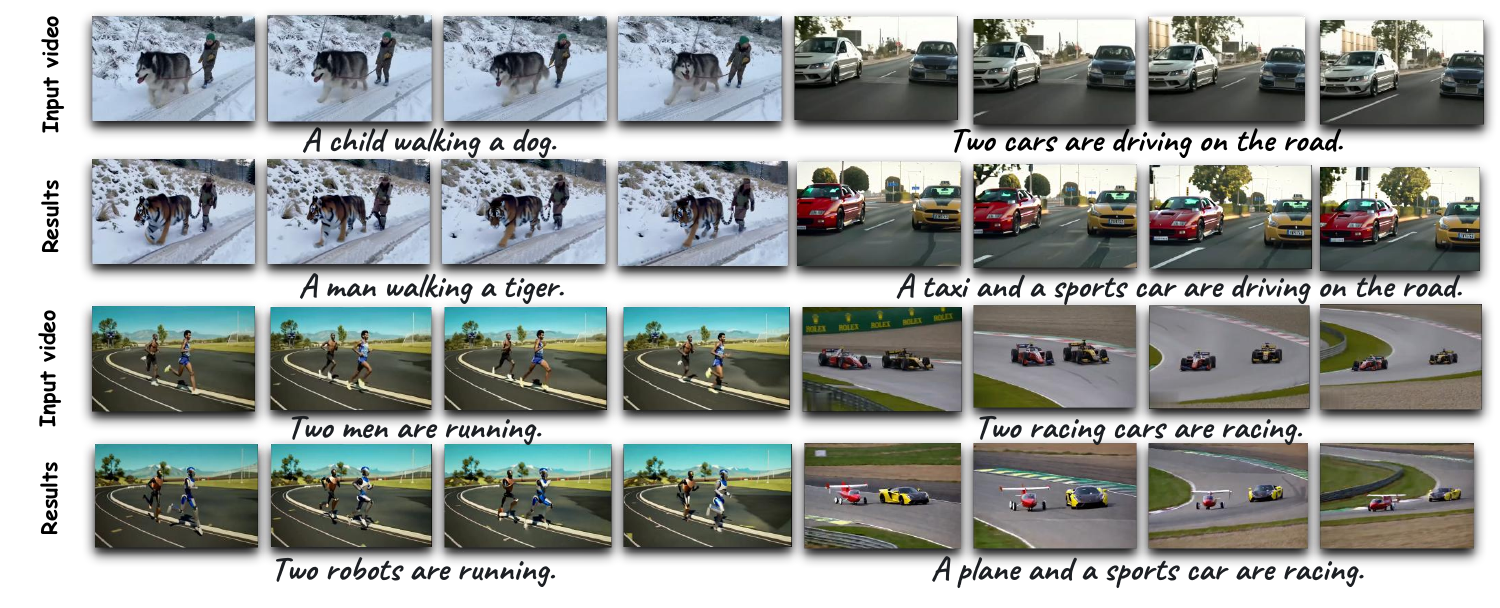}  
    \captionof{figure}{\textbf{Showcase of our MultiMotion.} Given an input video, MultiMotion can reproduce the same motion, capturing the dynamics of multiple moving objects.} 
    \label{fig:teaser}
}]

\begin{abstract}

Multi-object video motion transfer poses significant challenges for Diffusion Transformer (DiT) architectures due to inherent motion entanglement and lack of object-level control. We present MultiMotion, a novel unified framework that overcomes these limitations. Our core innovation is Mask-aware Attention Motion Flow (AMF), which utilizes SAM 2 masks to explicitly disentangle and control motion features for multiple objects within the DiT pipeline. Furthermore, we introduce RectPC, a high-order predictor-corrector solver for efficient and accurate sampling, particularly beneficial for multi-entity generation. To facilitate rigorous evaluation, we construct the first benchmark dataset specifically for DiT-based multi-object motion transfer. MultiMotion demonstrably achieves precise, semantically aligned, and temporally coherent motion transfer for multiple distinct objects, maintaining DiT's high quality and scalability.The code is in the supp.
\end{abstract}



\section{Introduction}

Imagine a world where virtual characters don't just exist in isolation, but can elegantly dance in perfect synchrony, where AI-driven animated ensembles fluidly interact, and where, in film effects, countless independent elements evolve with breathtaking realism. This is the grand vision of multi-object motion transfer – it's far more than simply replicating actions. It's about imbuing life and interaction into every single, independent entity within complex virtual scenes, as demonstrated in Fig.~\ref{fig:teaser}. Its applications are boundless, from precise virtual avatar control to large-scale multi-character animation. Yet, compared to the relative maturity of single-object motion transfer, the inherent complexities of multi-object scenarios, such as intricate motion disentanglement, precise semantic alignment, and the nuanced modeling of interactive behaviors, position it as a holy grail challenge in the pursuit of truly controllable video generation.
In recent years, diffusion models \citep{rombach2022high} have made remarkable strides in generating high-fidelity, temporally coherent video content. Among these, the Diffusion Transformer (DiT) \citep{peebles2023scalable} has rapidly emerged as a foundational element for many state-of-the-art systems, primarily due to its unified spatiotemporal attention mechanism and impressive scalability. However, it's precisely this global attention design that proves to be a fundamental limitation for DiT when confronted with multi-object scenarios. Lacking explicit awareness and separation of individual instances, DiT frequently succumbs to a pervasive and frustrating problem: motion entanglement. This means that in the latent space, the behaviors of different objects inevitably interfere with one another, leading to semantic drift, severely degraded object controllability, and ultimately, visually chaotic and incoherent outputs.

\begin{figure}[t]
\centering
\includegraphics[width=0.4\textwidth]{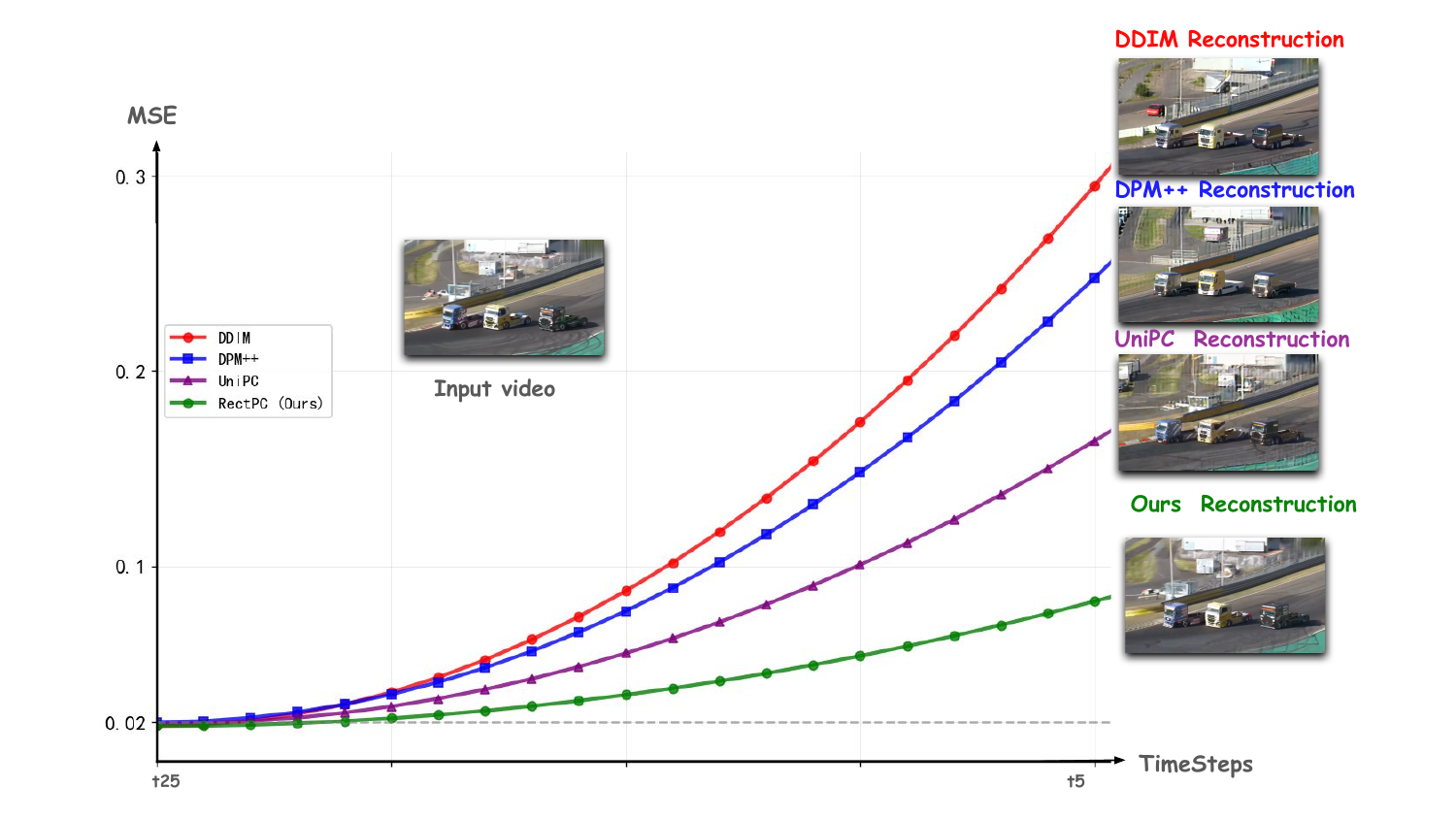}
\caption{
\textbf{Analysis of the inversion-reconstruction process. }
This figure shows the Mean Squared Error (MSE) between the intermediate latent representations from the inversion and reconstruction phases over $N$ timesteps. The curves represent the performance of different methods: DDIM (red), DPM++ (blue), UniPC (purple), and our proposed method, RectPC (green). The graph demonstrates that our method maintains significantly lower MSE throughout the process. The right side of the figure presents visual examples of the reconstructed images for each method, highlighting the superior fidelity of our RectPC method in comparison to the baselines.
}
\label{fig:rectpc}
\end{figure}

Moreover, existing mainstream diffusion inversion solvers, such as DDIM, DPM-Solver, and UniPC, are primarily designed and evaluated for the high-quality generation of single images or short, unconstrained video clips. While highly effective for their intended purposes, their fundamental design doesn't explicitly account for the intricate dynamics and inter-object consistency required in complex multi-object video editing tasks.  As shown in Fig.~\ref{fig:rectpc}, when directly applied to such scenarios, their limitations become evident: they often lead to high reconstruction errors, sluggish convergence, and pronounced instability, particularly under dynamic conditions involving occlusions or intricate interactions. Although some exploratory works attempt to infuse DiT with a degree of instance awareness through token masks or pose conditioning, these approaches largely remain confined to low-resolution generation or single-object animation. Crucially, they lack the necessary generality and robustness required for scalable, high-fidelity multi-object motion control.

To directly and comprehensively address these persistent challenges plaguing the field of multi-object motion transfer, we proudly introduce MultiMotion – a unified and pioneering framework meticulously engineered for multi-object motion transfer within DiT architectures. MultiMotion, in an unprecedented manner, achieves precise object-specific motion representation, controllable attention disentanglement, and efficient high-order diffusion inversion, all seamlessly integrated within a single, coherent pipeline. Our methodology is built upon two disruptive innovations: first, we enhance the concept of Attention Motion Flow by introducing a novel Mask-aware Attention Motion Flow (AMF) mechanism, which ingeniously leverages SAM 2's \citep{ravisam} precise instance-level masks to fundamentally disentangle and inject instance-aware attention, enabling fine-grained control over multi-object behaviors; second, our advanced predictor-corrector solver, RectPC. While RectPC is specifically designed to ensure the stability and exceptional precision of complex multi-entity generation, its core architectural principles also confer a strong generalization capability, allowing its benefits to extend to a wider range of diffusion modeling tasks.

To comprehensively validate our method's effectiveness and address the evaluation gap in this domain, we construct the first benchmark dataset specifically for multi-object motion transfer—MultiMotionEval. This dataset comprises 103 high-quality videos, each with 41 frames and a resolution of 832×480. It systematically captures diverse multi-object dynamics and complex interactions, featuring 321 distinct objects. Over  80\% of the scenes  include two or more objects, involving challenging interactive modes such as chasing, occlusion, synchronous collaboration, and separation. We provide detailed instance-level masks and trajectory annotations for all videos. MultiMotionEval is a high-value and challenging benchmark that provides an indispensable resource for the rigorous, quantitative evaluation of a model's object-level controllability, temporal consistency, and robustness under complex interactions.

Through these innovations, our work makes the following key contributions:
\begin{itemize}
    \item We propose MultiMotion, the first unified framework for disentangled multi-object motion transfer in DiT. It introduces Mask-aware Attention Motion Flow (AMF), building upon prior AMF concepts by incorporating SAM 2 masks for precise object-level motion disentanglement and control.
    
    \item We develop RectPC, a high-order predictor-corrector solver formulated in the reparameterized $\lambda$-space. RectPC combines extrapolation, finite-difference correction, and midpoint refinement to enable efficient and stable sampling with significantly fewer steps.
    
    \item To comprehensively validate effectiveness and address the evaluation gap in this domain, we construct the MultiMotionEval, the first benchmark dataset specifically for multi-object motion transfer, comprising 103 high-quality videos. We show extensive evaluations on MultiMotionEval to verify the superiority of MultiMotion.
\end{itemize}

\begin{figure*}[t]
\centering
\includegraphics[width=0.9\textwidth]{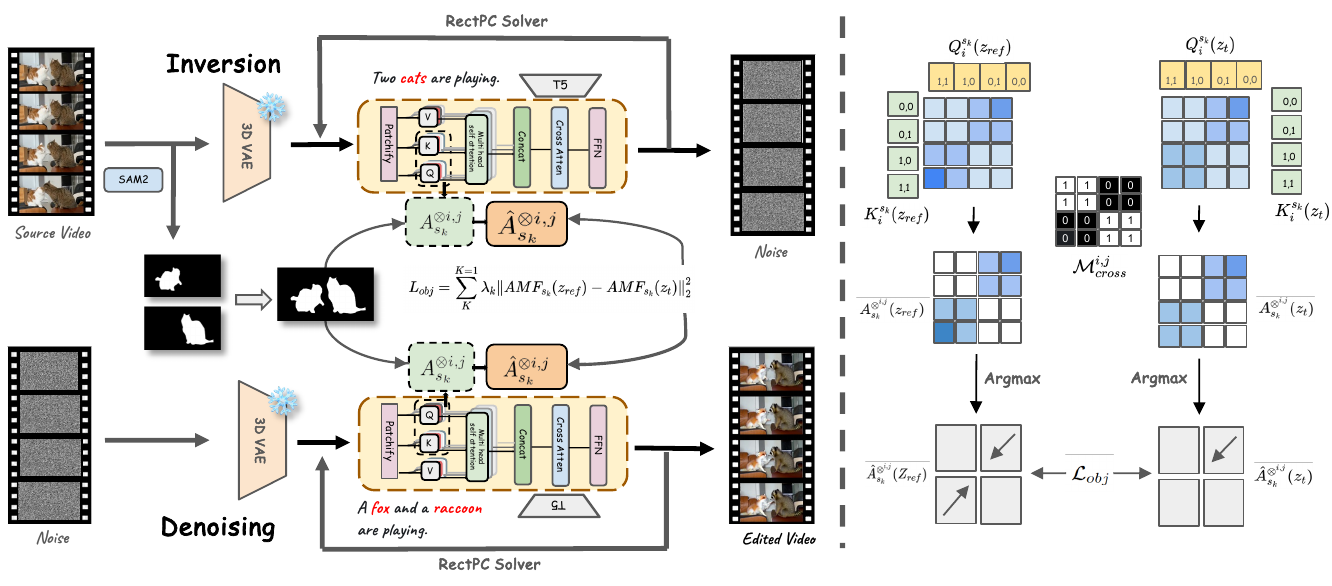}
\caption{
\textbf{The overview of the MultiMotion.}
The source video is encoded by a 3D VAE and processed by SAM2 to obtain instance-level masks.
Based on these masks, we extract object-specific motion fields via Mask-aware Attention Motion Flow (AMF).
During generation, \textit{RectPC Solver} iteratively refines the trajectory via high-order diffusion inversion in $\lambda$-space.
This enables accurate and controllable multi-object motion transfer in the edited video.
}
\label{fig:framework}
\end{figure*}

\section{Related Work}

\subsubsection{Video Motion Transfer}
Motion transfer focuses on synthesizing new video sequences that preserve motion dynamics from a reference video. Traditional approaches \citep{guo2023animatediff,xing2024make,xing2024dynamicrafter,zhang2025framepainter}  rely on explicit control signals like pose, flow, or segmentation masks, often demanding extensive annotated datasets and significant computational resources. More recent work explores implicit motion control, either in a training-free manner\citep{hu2024motionmaster,xu2025hunyuanportrait,ma2024followpose, ma2025followfaster, ma2025controllable, ma2025followcreation, ma2025followyourmotion, ma2024followyouremoji, ma2025followyourclick, ma2023magicstick, ma2022visual, pondaven2025video,yesiltepe2024motionshop,long2025follow}—where motion embeddings guide generation via gradients during inference—or through tuning-based paradigms \citep{jeong2024dreammotion,yan2025eedit, zhang2025zero, zhang2025magiccolor, zhu2024instantswap, wang2024cove, feng2025dit4edit, chen2024follow, feng2025follow, yuluo2025gr, wan2025unipaint, chen2025contextflow, long2025follow, shen2025follow, zhao2024motiondirector}using parameter-efficient modules such as LoRA to decouple motion and appearance. However, most of these methods are primarily designed for UNet-based models, and their applicability to transformer-based architectures, particularly DiT, remains limited. Crucially, they often struggle with fine-grained multi-object control, especially in complex scenes involving interactions and occlusions.
\subsubsection{Inversion}
Inversion in diffusion models aims to recover the latent noise representation from real visual data by reversing the generation trajectory. A foundational technique, DDIM inversion\citep{songdenoising,songscore}, recursively adds predicted noise across forward steps to approximate this trajectory. However, discretization errors in this process can degrade reconstruction fidelity, particularly for long sequences or intricate motion content. To mitigate this, several methods\citep{elarabawy2022direct,song2025idprotector, song2024anti,hui2025autoregressive, ci2024wmadapter,lu2025dpm,miyake2025negative,mokady2023null,rout2024beyond,wallace2023edict} have introduced high-order solvers or prediction-correction mechanisms to improve stability and and accuracy.
More recent approaches, like \citep{routsemantic,wangtaming}  and UniPC-Solver \citep{zhaounipc}, offer stronger consistency and adaptive control. RF-Solver improves precision through high-order modeling and history reuse, while UniPC-Solver focuses on efficient inference via linear path prediction and direct sampling. Nevertheless, a persistent challenge remains in balancing accuracy and speed across diverse diffusion tasks, particularly in complex video scenarios involving multiple dynamic entities or subtle temporal dependencies. Overcoming these limitations in the context of high-fidelity, controllable multi-object video generation represents a critical unmet need.

\section{Method}

Given a reference video, video motion transfer aims to  synthesize the video with same object motions and camera pose. The pipeline of our MultiMotion is shown in Fig.~\ref{fig:framework}. In the following section, we first introduce the Multi-Object Motion Decomposition in Sec. 3.1.  The Mask-aware Attention Motion Flow is present in Sec. 3.2. Then the Multi-Object Motion Recomposition is following in Sec. 3.3.  Finally, we demonstrate the RectPC Solver in Sec. 3.4.

\subsection{Multi-Object Motion Decomposition}
\label{sec:Multi-Object Motion Decomposition}

\subsubsection{Instance-Level Semantic Segmentation}
To robustly handle multi-object motion transfer, our framework first performs instance-level semantic segmentation and then decouples motion regions.
Given a reference video $V_{ref} = \{x_1, x_2, \ldots, x_F\}$, we utilize the powerful SAM 2 model to obtain precise trajectory masks for each object. For the $k$-th object $s_k$, SAM 2 generates its mask sequence across all frames $\mathcal{M}_{s_k} = \{M_{s_k}^1, M_{s_k}^2, \ldots, M_{s_k}^F\}$, where $M_{s_k}^i \in \{0,1\}^{H \times W}$ represents the binary mask of object $s_k$ in the $i$-th frame.

\subsubsection{Motion Region Decoupling}
To prevent motion confusion and ensure true independence in multi-object scenarios, we propose a refined motion region decoupling strategy. For object $s_k$'s motion between frames $i$ and $j$, its independent motion region is defined as:
\begin{equation}
M_{s_k}^{i|j} = M_{s_k}^i \setminus \bigcup_{m \neq k} (M_{s_k}^i \cap M_{s_m}^j)
\end{equation}
where $\setminus$ denotes the set difference operation. This operation ensures that object $s_k$'s motion features are not interfered with by other objects' motions, which is critical for accurate motion decoupling.

\subsection{Mask-aware Attention Motion Flow}
\label{sec:Mask-aware Attention Motion Flow}
Building upon the DiT architecture, we introduce Mask-aware Attention Motion Flow (AMF) to enable fine-grained, object-specific motion guidance, directly addressing the limitations of global attention in multi-object settings.

\subsubsection{DiT Attention Mechanism Analysis}
For the $n$-th layer of the DiT model, we analyze its self-attention features at denoising step $t=0$. Given the latent representation $z_{ref} = \mathcal{E}(V_{ref})$, the DiT block computes query and key matrices as part of its denoising process:
\begin{equation}
\{Q, K\}^n \leftarrow \epsilon_\theta(z_{ref}, \emptyset, 0, \rho)
\end{equation}
where $\rho$ is the positional embedding and $\epsilon_\theta$ is the DiT denoising network.

\subsubsection{Mask-Guided Cross-Frame Attention}
Traditional AMF methods compute global cross-frame attention, which, as discussed, easily leads to motion confusion in multi-object scenarios. To mitigate this, we propose mask-guided cross-frame attention computation:
\begin{equation}
A_{s_k}^{\otimes i,j} = \sigma\left(\tau \frac{Q_i^{s_k} (K_j^{s_k})^T}{\sqrt{d_k}}\right) \odot \mathcal{M}_{cross}^{i,j}
\end{equation}
Here:
\begin{itemize}
    \item $Q_i^{s_k}$ and $K_j^{s_k}$ represent the query and key features derived from object $s_k$'s region in frames $i$ and $j$, respectively.
    \item $\mathcal{M}_{cross}^{i,j}$ is a cross-frame mask constraint matrix, meticulously constructed to ensure attention calculation only occurs within valid and decoupled object regions, preventing inter-object interference.
    \item $\odot$ denotes element-wise multiplication.
\end{itemize}

\subsubsection{Object-Specific Motion Flow Construction}
Based on this mask-guided attention, we construct independent motion flows for each object. For object $s_k$, we first obtain the strongest attention correspondences by applying an argmax operation:
\begin{equation}
\hat{A}_{s_k}^{\otimes i,j} = \text{argmax}(A_{s_k}^{\otimes i,j})
\end{equation}
Then, we construct the object-specific displacement matrix $\Delta_{s_k}^{i,j}$, where each element represents a patch's motion vector from frame $i$ to frame $j$. Finally, object $s_k$'s attention motion flow (AMF) is defined as the collection of these displacement matrices:
\begin{equation}
\text{AMF}_{s_k}(z_{ref}) = \{\Delta_{s_k}^{i,j}\}_{i,j \in [1,F]}
\end{equation}

\subsection{Multi-Object Motion Recomposition}
\label{sec:Multi-Object Motion Recomposition}
With the object-specific AMF extracted, we guide the target video generation process to ensure accurate and disentangled motion transfer for all entities.

\subsubsection{Object-Specific Motion Guidance}
During target video generation, we use the extracted object-specific AMF as precise guidance signals. For each object $s_k$, we compute the current soft motion flow at denoising step $t$:
\begin{equation}
\tilde{\Delta}_{s_k}^{i,j}(t) = \sum_{p} A_{s_k}^{\otimes i,j}(p) \cdot \text{pos}(p)
\end{equation}

\subsubsection{Multi-Object Motion Loss Function}
To enforce adherence to the desired object-specific motions and maintain background consistency, we define a comprehensive multi-object motion loss:
\begin{equation}
\mathcal{L}_{obj} = \sum_{k=1}^K \lambda_k \left\|\text{AMF}_{s_k}(z_{ref}) - \text{AMF}_{s_k}(z_t)\right\|_2^2
\end{equation}
\begin{equation}
\mathcal{L}_{bg} = \lambda_c \left\|\text{AMF}_{bg}(z_{ref}) - \text{AMF}_{bg}(z_t)\right\|_2^2
\end{equation}
The total multi-object loss is then:
\begin{equation}
\mathcal{L}_{multi} = \mathcal{L}_{obj} + \mathcal{L}_{bg}
\end{equation}

\subsubsection{Adaptive Weight Adjustment}
To robustly handle complex interactions and occlusions between objects, we introduce an adaptive weight adjustment mechanism for each object's loss contribution:
\begin{equation}
\lambda_k^{adaptive} = \lambda_k \cdot \exp\left(-\alpha \cdot \text{IoU}\left(M_{s_k}^i, \bigcup_{m \neq k} M_{s_m}^j\right)\right)
\end{equation}
This adaptive weighting dynamically reduces an object's motion loss influence when it's heavily occluded by other objects, preventing erroneous guidance signals in ambiguous regions.

\subsection{RectPC Solver}
\label{sec:RectPC Solver}
To further enhance the sampling accuracy and efficiency of diffusion models, especially crucial for multi-object video generation, we propose RectPC, a high-order predictor-corrector (PC) solver formulated in the reparameterized $\lambda$-space. This solver provides stable and accurate trajectory updates without modifying the underlying model architecture, directly addressing the limitations of traditional solvers in complex tasks.

\subsubsection{Reparameterization in $\lambda$-Space}
We adopt the $\lambda$-space reparameterization, where $\lambda_t = \log \alpha_t - \log \sigma_t$. This transformation provides a more stable and linear path for numerical integration, facilitating high-order predictions.

\subsubsection{High-Order Extrapolation Estimator}
Given historical noise predictions $\{\hat{\epsilon}_{s_0}, \dots, \hat{\epsilon}_{s_{K-1}}\}$ from previous steps, our high-order extrapolation estimator predicts the next state:
\begin{equation}
\mathbf{x}_{\lambda_t}^{\text{pred}} = A \mathbf{x}_{\lambda_{t-1}} - B \cdot \phi_1(h) \cdot \hat{\epsilon}_{s_0} - B \sum_{i=1}^{K-1} \rho_i D_i
\end{equation}
where $D_i = \hat{\epsilon}_{s_i} - \hat{\epsilon}_{s_{i-1}}$ represents the difference in historical noise estimates, and $\rho_i$ are weights meticulously solved via a Vandermonde system to ensure high-order accuracy.

\subsubsection{Midpoint Correction}
Optionally, we refine the predicted state with a midpoint correction step, which significantly improves trajectory accuracy and stability:
\begin{equation}
\mathbf{x}_{\lambda_t}^{\text{corr}} = \mathbf{x}_{\lambda_t}^{\text{pred}} + \frac{h^2}{2} \cdot \left( \frac{v_\theta(\mathbf{x}_{\text{mid}}, t) - v_\theta(\mathbf{x}_{\lambda_t}^{\text{pred}}, t)}{h} \right)
\end{equation}
This correction term dynamically adjusts the trajectory based on the model's prediction at the midpoint, effectively reducing cumulative errors.

\subsubsection{Inference Procedure}
The overall RectPC inference procedure is detailed in Algorithm \ref{alg:rectpc}. It leverages the high-order estimator and optional midpoint correction to iteratively refine the latent representation from noise to data.

\begin{algorithm}[t]
\caption{RectPC Inference Procedure}
\label{alg:rectpc}
\textbf{Input}: Initial Gaussian noise $\mathbf{x}_{\lambda_T}$, timestep list $\{\lambda_t\}_{t=0}^T$, model $v_\theta$ \\
\mbox{\textbf{Output:} Reconstructed sample $\mathbf{x}_{\lambda_0}$}
\begin{algorithmic}[1]
\FOR{$t = T$ down to $1$}
    \STATE Predict: $\hat{\epsilon}_{s_0} = v_\theta(\mathbf{x}_{\lambda_t}, t)$
    \STATE Retrieve past estimates $\{\hat{\epsilon}_{s_1}, \dots, \hat{\epsilon}_{s_{K-1}}\}$ (if $K > 1$)
    \STATE Compute extrapolated prediction: $\mathbf{x}_{\lambda_{t-1}}^{\text{pred}}$ using high-order estimator
    \IF{midpoint correction enabled}
        \STATE Compute midpoint: $\mathbf{x}_{\text{mid}} = \frac{1}{2}(\mathbf{x}_{\lambda_t} + \mathbf{x}_{\lambda_{t-1}}^{\text{pred}})$
        \STATE Compute corrected state: $\mathbf{x}_{\lambda_{t-1}}^{\text{corr}}$
    \ELSE
        \STATE $\mathbf{x}_{\lambda_{t-1}}^{\text{corr}} \gets \mathbf{x}_{\lambda_{t-1}}^{\text{pred}}$
    \ENDIF
    \STATE Update: $\mathbf{x}_{\lambda_{t-1}} \gets \mathbf{x}_{\lambda_{t-1}}^{\text{corr}}$
\ENDFOR
\STATE \textbf{return} $\mathbf{x}_{\lambda_0}$
\end{algorithmic}
\end{algorithm}

\subsection{Overall Workflow}
The complete MultiMotion generation process, integrating Mask-aware AMF guidance with the RectPC solver, is outlined in Algorithm \ref{alg:multimotion}.

Initially, for a given reference video $V_{ref}$, we encode it into latent space $z_{ref}$ using a 3D VAE. Concurrently, SAM 2 processes $V_{ref}$ to provide precise instance-level masks for all objects. These masks are then used to compute refined, decoupled motion regions. Utilizing these decoupled regions, we extract object-specific AMF features, $\text{AMF}_{s_k}(z_{ref})$, which capture the desired motion dynamics for each individual object.

During the iterative denoising process, starting from pure Gaussian noise $z_T$, at each step $t$:
\begin{enumerate}
    \item The DiT model produces a denoised latent $z_t$.
    \item The current AMF for each object, $\text{AMF}_{s_k}(z_t)$, is computed from $z_t$.
    \item A comprehensive multi-object loss $\mathcal{L}_{multi}$ is calculated by comparing the current AMF with the reference AMF, incorporating adaptive weighting for robustness against occlusions.
    \item The RectPC solver then updates the latent representation $z_{t-1}$, leveraging its high-order prediction and correction mechanisms, guided by the combined $\mathcal{L}_{multi}$ to precisely steer the multi-object motion.
\end{enumerate}
This iterative process continues until the final denoised latent $z_0$ is obtained, which is then decoded by the 3D VAE to yield the target video $V_{target}$ with accurate and controllable multi-object motion.

\begin{algorithm}[t]
\caption{MultiMotion Multi-Object Motion Transfer}
\label{alg:multimotion}
\textbf{Input}: Reference video $V_{ref}$, Target condition $C_{target}$\\
\textbf{Output}: Target video $V_{target}$
\begin{algorithmic}[1]
\STATE Encode $V_{ref}$ to $z_{ref}$ using 3D VAE
\STATE Use SAM 2 to obtain multi-object masks $\{M_{s_k}\}$ from $V_{ref}$
\STATE Compute refined motion regions $\{M_{s_k}^{i|j}\}$ for each object
\STATE Extract object-specific AMF from $z_{ref}$: $\{\text{AMF}_{s_k}(z_{ref})\}$
\STATE Initialize: $z_T \sim \mathcal{N}(0,I)$
\FOR{$t = T$ down to $1$}
\STATE Compute current AMF from $z_t$: $\{\text{AMF}_{s_k}(z_t)\}$
\STATE Compute multi-object loss: $\mathcal{L}_{multi}$ (with adaptive weights)
\STATE Update latent using RectPC: $z_{t-1} = \text{RectPC\_update}(z_t, \text{current\_timestep}, \mathcal{L}_{multi})$
\ENDFOR
\STATE \textbf{return} $V_{target} = \mathcal{D}(z_0)$ (decode using 3D VAE)
\end{algorithmic}
\end{algorithm}

\begin{figure*}[t]
\centering
\includegraphics[width=0.8\linewidth]{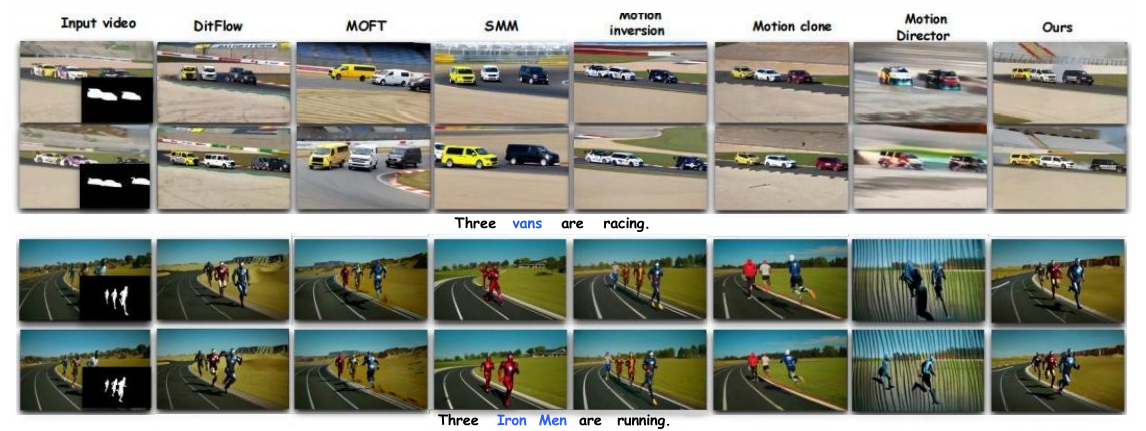}
\caption{\textbf{Qualitative comparison with baselines.} We conduct visual comparisons with six baseline methods across a variety of motion types, especially those involving multiple objects. More comparisions can be found in supplementary}
\label{fig:comparison_1}
\end{figure*}
\section{Experiments}
\subsection{Implementation details}

\begin{figure*}[t]
\centering
\includegraphics[width=\linewidth]{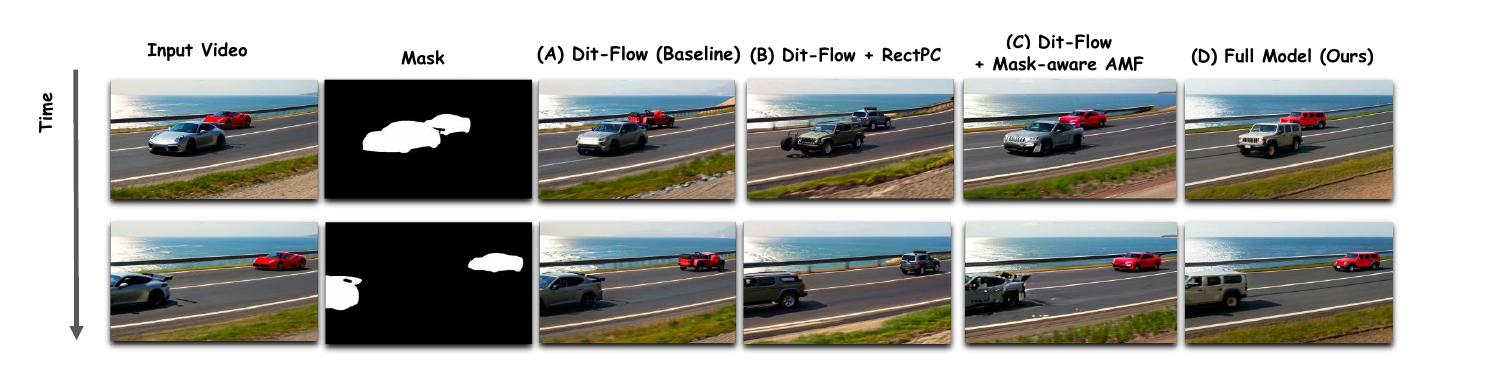}
\caption{\textbf{Ablation study about proposed modules}. We systematically evaluate the effectiveness of our proposed modules. The prompt used for generation is "Two cars are driving on the road." The figure displays two consecutive frames (top and bottom rows) for each model variant: (A) The Dit-Flow baseline with its original solver and AMF exhibits motion blur and entanglement. (B) The variant with our RectPC solver improves motion fidelity and temporal consistency. (C) The variant with Mask-aware AMF shows better disentanglement of the two cars' movements and improved adherence to object masks. (D) Our full model, which combines the RectPC solver and the Mask-aware AMF, achieves superior performance by correctly reconstructing all subjects with precise, disentangled motion.The prompt is "Two jeeps are driving on the road".}
\label{fig:ablation}
\end{figure*}

We adopt a uniform denoising process of 70 steps across all baseline methods. During the first 20 denoising timesteps, we perform 5 steps of fine-tuning using the Adam optimizer \citep{kingma2014adam}, with a linearly decaying learning rate from 0.008 to 0.002, following the optimization strategy outlined in \citep{yatim2024space}. For the computation of the AMF loss, we select the 15th Transformer block of the WAN2.1-1.3B model as the evaluation layer for feature alignment.
More details can be found in supplementary.

\subsection{MultiMotionEval}
To address the lack of standardized benchmarks in the field of multi-object motion transfer, we introduce MultiMotionEval, a dedicated evaluation suite designed to assess the capabilities of motion transfer methods involving multiple entities. The dataset comprises 103 video clips featuring diverse multi-object motion scenarios. Specifically, single-object motion sequences focus on a variety of movement patterns performed by a single subject, while multi-object motion emphasizes the spatial relationship consistency and coordinated behavior among multiple instances.

All videos are sourced from publicly licensed video platforms, and caption annotations are automatically generated using GPT4o. Each video lasts approximately 2 seconds, consisting of 41 frames—making it suitable for short-range motion transfer evaluation. MultiMotionEval offers a standardized evaluation protocol that spans various motion categories, enabling systematic and fair comparison of motion transfer methods from multiple perspectives, including semantic consistency, temporal coherence, and spatial alignment. This benchmark fills an important gap in the current landscape of multi-object motion transfer evaluation.

\subsection{Comparison with Existing Methods}

We conducted a systematic comparison with current state-of-the-art video motion transfer methods, including MOFT\citep{xiao2024video}, MotionInversion \citep{wang2025motion}, MotionClone \citep{lingmotionclone}, SMM, MotionDirector, and DiTFlow. Motionshop and MotionCrafter \citep{zhang2023motioncrafter} were excluded from the comparison due to the lack of public releases. Experimental results demonstrate that our proposed method, MultiMotion, achieves superior generation quality and greater robustness across various motion types. In single-object motion transfer tasks, existing methods often fail to accurately replicate the motion trajectory of the reference video, resulting in broken or misaligned action rhythms. In contrast, our method precisely extracts motion patterns from the reference and ensures smooth, natural motion in the generated video. In multi-object motion scenarios, models such as MotionDirector and SMM struggle to preserve spatial relationships and synchronized movement among multiple objects. Our model successfully maintains inter-object spatial consistency and coordinated motion dynamics. For camera motion transfer, our method also delivers better continuity and stability in viewpoint transitions, outperforming other methods in overall visual quality.
\begin{table}[t]
\centering
\caption{\textbf{Comparison with state-of-the-art video motion transfer methods}.}
\label{tab:comparison_main}
\scalebox{0.8}{
\begin{tabular}{l@{\hskip 0.5cm}ccc}
\toprule
\hline
\textbf{Method} & \textbf{Text Sim.}$\uparrow$ & \textbf{Motion Fid.}$\uparrow$ & \textbf{Temp. Cons.}$\uparrow$ \\
\hline
MOFT & 0.290 & 0.795 & 0.935 \\
MotionClone & 0.305 & 0.835 & 0.912 \\
SMM & 0.280 & 0.920 & 0.930 \\
DiTFlow & 0.368 & 0.820 & 0.940 \\
MotionInversion & 0.308 & 0.845 & 0.775 \\
MotionDirector & 0.292 & 0.910 & 0.950 \\
\textbf{Ours} & \textbf{0.385} & \textbf{0.985} & \textbf{0.978} \\
\hline
\bottomrule
\end{tabular}
}
\end{table}

For quantitative evaluation, we tested all models on our self-constructed MultiMotionEval benchmark under the same settings, using videos with forty-one frames and a resolution of eight hundred thirty-two by four hundred eighty pixels. Evaluation metrics include motion fidelity, which measures the similarity of object trajectories between reference and generated videos; temporal consistency, which uses CLIP feature similarity between consecutive frames to assess coherence; and text alignment, which evaluates semantic consistency through the average cosine similarity between extracted video features and the input text prompt. 
Fig.~\ref{fig:comparison_1} provides qualitative comparisons demonstrating the superior visual quality of our method across various motion scenarios. The quantitative results in Table~\ref{tab:comparison_main} further validate our method's effectiveness, showing significant improvements across all evaluation metrics.
\subsection{Ablation Study}
To thoroughly understand the contributions of each component within our proposed MultiMotion framework, we conducted a series of rigorous ablation experiments. Our research focuses on two key innovations: the RectPC solver and the Mask-aware Attention Motion Flow (Mask-aware AMF). To ensure fairness and compelling evidence, we selected Dit-Flow as our baseline model, as it is a representative state-of-the-art method in this domain. By systematically removing or replacing key components of the framework, we quantified their impact on the multi-subject video motion transfer task.

All experiments were conducted under identical configurations and on the same dataset. We used the following key metrics for evaluation: Text Similarity (Text Sim.), to measure the alignment between the generated content and the textual description; Motion Fidelity (Motion Fid.), to assess the realism and accuracy of the generated motion; and Temporal Consistency (Temp. Cons.), to evaluate the smoothness and coherence of motion between video frames.

\begin{table}[t]
\centering
\caption{\textbf{Ablation study on the core components of the MultiMotion framework.} This table demonstrates the incremental contribution of the RectPC solver and Mask-aware AMF to the overall performance.}
\label{tab:ablation_study}
\scalebox{0.7}{
\begin{tabular}{l@{\hskip 0.5cm}ccc}
\toprule
\hline
\textbf{Model Variant} & \textbf{Text Sim.}$\uparrow$ & \textbf{Motion Fid.}$\uparrow$ & \textbf{Temp. Cons.}$\uparrow$ \\
\hline
(A) DiT-Flow (Baseline) & 0.368 & 0.820 & 0.940 \\
(B) DiT-Flow + RectPC & 0.375 & 0.900 & 0.965 \\
(C) DiT-Flow + Mask-aware AMF & 0.378 & 0.850 & 0.945 \\
(D) Full Model (Ours) & \textbf{0.385} & \textbf{0.985} & \textbf{0.978} \\
\hline
\bottomrule
\end{tabular}
}
\end{table}
By comparing the Dit-Flow baseline model (A) with the variant that introduces the RectPC solver (B), we observe a significant improvement in both Motion Fidelity and Temporal Consistency while all other components remain unchanged. This provides strong evidence for the superiority of the RectPC solver in achieving efficient sampling and stable video generation, laying a solid foundation for precise motion control. We further validate the effectiveness of our attention flow design by comparing the baseline model (A) with the variant using the Mask-aware AMF (C). The results show that the introduction of Mask-aware AMF substantially boosts Text Similarity and enhances Motion Fidelity. This indicates that leveraging object masks for explicit control successfully disentangles the motion of multiple subjects, effectively mitigating the common issue of motion entanglement in existing DiT architectures. Our full model (D), which integrates both the RectPC solver and Mask-aware AMF into the Dit-Flow baseline, achieves the best performance across all evaluation metrics. Compared to variants (B) and (C), this further demonstrates that our two innovative components do not operate independently but synergistically, collaboratively achieving high-quality, high-fidelity, and semantically accurate multi-subject motion transfer. This result underscores the rationality and effectiveness of our unified framework design.The results of our comprehensive ablation study are presented in Table~\ref{tab:ablation_study} and visualized in Fig.~\ref{fig:ablation}.

\section{Discussion and Conclusion}
In this work, we propose MultiMotion, a novel multi-object motion transfer framework tailored for Diffusion Transformer (DiT) architectures. To tackle the challenges of motion ambiguity and semantic entanglement, we introduce the Mask-aware Attention Motion Flow (AMF) mechanism, enabling instance-level motion disentanglement and precise motion feature extraction. Furthermore, we incorporate the high-order RectPC sampling strategy with midpoint correction and extrapolation to improve inversion efficiency and stability. To comprehensively validate effectiveness and address the evaluation gap in this domain, we construct the MultiMotionEval, the first benchmark dataset specifically for multi-object motion transfer, comprising 103 high-quality videos. We show extensive evaluations on Multi-MotionEval to verify the superiority of MultiMotion.
We believe MultiMotion not only advances the frontier of controllable video generation but also provides a generalizable framework for future research in fine-grained video editing and diffusion model inversion.

\bibliography{aaai2026}
\subfile{supp}

\end{document}

%% file: supp.tex
\clearpage

\section{Supplementary Material}

\section{Overview}
In the supplementary material, we provide additional content and experimental results related to our work. First, we include related work that was not covered in the main text, particularly studies related to T2V (Sec.2). Next, we present a gallery of project visualizations to more intuitively demonstrate the qualitative performance of our method (Sec.3). In Sec.4, we offer additional comparative examples to further validate the advantages of our method over baselines. Finally, we discuss the limitations of our method regarding the use of masks (Sec.5). In addition, the supplementary material also includes sample videos and source code for further reference.

\section{Text-to-Video Generation}
Text-to-video generation aims to synthesize realistic and coherent video sequences from descriptive language prompts. Early methods  \citep{guo2023animatediff,zhang2025easycontrol, zhang2024ssr, song2025layertracer, song2025makeanything, ma2025followyourmotion, huang2025photodoodle,ma2022visual, Z1, Z2, Z3, Z4, Z5, MAT, Z6, Z7, Z8, Z9, zhang2025follow,zhao2023controlvideo,zhang2025magiccolor,zhu2025multibooth,liu2025avatarartist,yan2025eedit,yang2023uni,chen2024follow,wangtaming,yatim2024space} augmented temporal modules within UNet-based diffusion architectures to enhance temporal consistency. More recently, large-scale pretrained models based on Diffusion Transformers (DiTs) have demonstrated superior generation quality and scalability. Representative examples include Sora \citep{liu2024sora}, CogVideoX \citep{yangcogvideox}, EasyAnimate \citep{xu2024easyanimate}, HunyuanVideo \citep{kong2024hunyuanvideo}, and wan \citep{wan2025wan}, all capable of generating high-fidelity video from detailed text prompts. These advancements have significantly improved spatial-temporal alignment, opening new avenues for controllable video synthesis.

\paragraph{Related work}
We thanks these related work~\cite{sun2025gl, sun2025bs, zhang2025echomask, zhang2025semtalk, wang2025characterfactory, liu2023delving, liu2024headartist, liu2023human} and their contribution. Also, we are motivated by some related works~\cite{zhao2025unified, Zhao_2023_CVPR, Zhao_2023_ICCV_DDFM, zhu2022one, feng20254dgs, li2025gs2e, shenlong, shen2024imagpose, shen2025imagdressing, shen2024advancing, shen2025imaggarment, shen2025imagedit, shen2025imagharmony, jia2019comdefend, jia2022adversarial, jia2022boosting,Z1,Z2,Z3,Z4,Z5, MAT,Z6,Z7,Z8,Z9,she2025customvideox,wang2025mint,ying2024restorerid,liu2024llm4gen,liu2024llm4gen,shen2025follow, wan2025unipaint}. Additionally, we also care about some AI-safety works~\cite{song2025idprotector,song2024anti,hui2025autoregressive,ci2024wmadapter,ci2024ringid,liu2024image,yang2024can}

\section{Project Gallery}
\begin{figure*}[t]
\centering
\includegraphics[width=\linewidth]{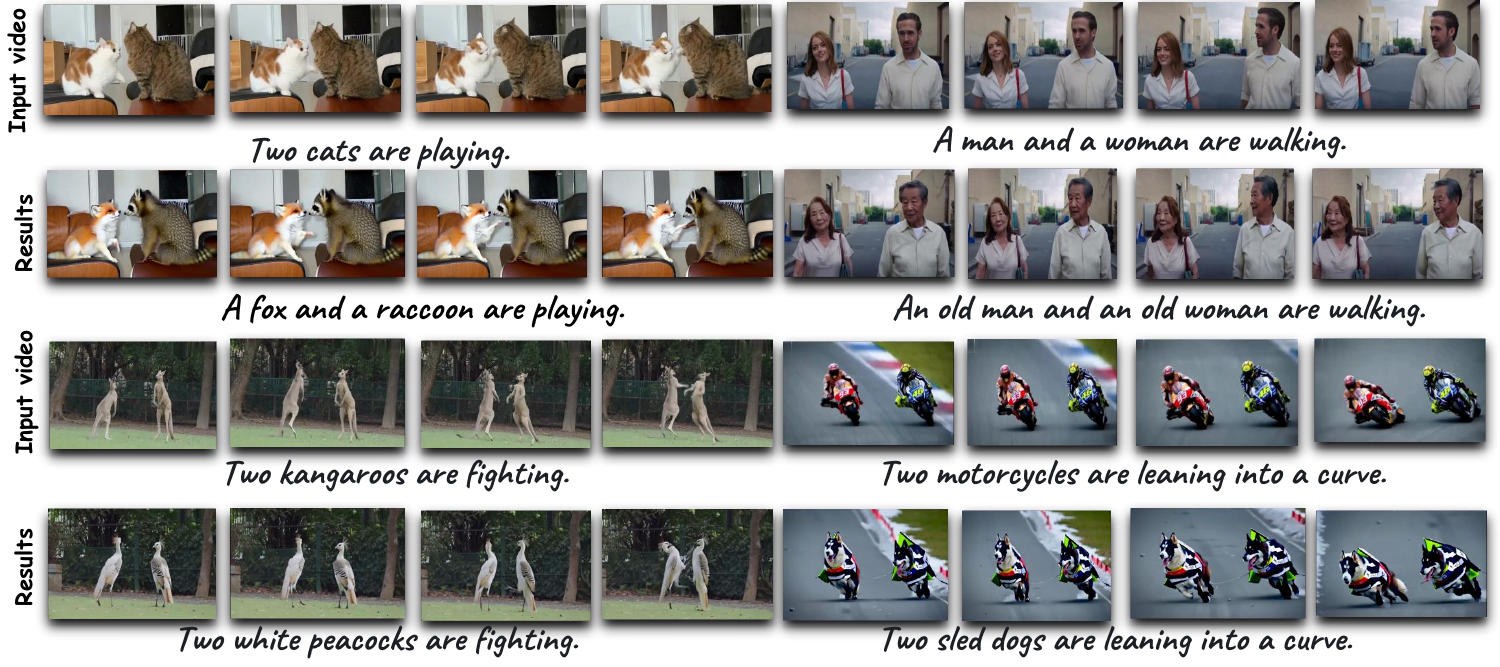}
\caption{\textbf{Gallery of our proposed method. Given a reference video, our MultiMotion model can generate a high-quality video clip that replicates the motion of multiple objects.}}
\label{fig:gallery}
\end{figure*}

To better illustrate the qualitative performance of our method, this section presents a set of representative visual results covering diverse input scenarios, demonstrating the generalization ability and robustness of our approach under various conditions.
As shown in Figure~\ref{fig:gallery}, we showcase examples involving different object categories, lighting conditions, background environments, and appearance variations, highlighting the method’s adaptability to complex visual inputs.
\section{Additional Comparative Results}
To further validate the effectiveness of our method, this section presents additional comparative experiments with several baseline approaches, aiming to highlight the advantages of our method under varying settings and input conditions.

\begin{figure*}[t]
\centering
\includegraphics[width=\linewidth]{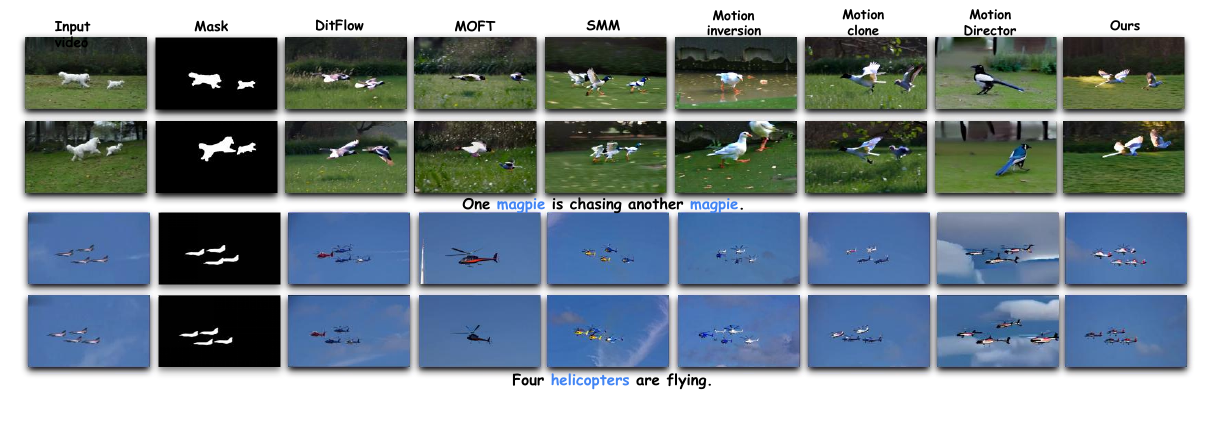}
\caption{\textbf{Qualitative comparison with baselines.} We conduct visual comparisons with six baseline methods across a variety of motion types, especially those involving multiple objects.}
\label{fig:comparison}
\end{figure*}

Figure~\ref{fig:comparison} presents qualitative comparisons with six baseline methods across a variety of motion types, especially those involving multiple objects. The results demonstrate our method's superior performance in preserving motion fidelity and temporal consistency.
\section{Limitation on Mask Usage}
While our method demonstrates strong performance across a variety of tasks, it still relies on accurate mask guidance to produce optimal results. Specifically, our framework assumes that the provided masks are spatially aligned with the relevant regions of interest. When masks are noisy, imprecise, or poorly aligned—such as those generated by weak segmentation models or in highly cluttered scenes—the quality of the output may degrade significantly.

In particular, inaccurate masks can lead to issues such as bleeding artifacts, incomplete transfers, or spatial distortions in the generated outputs. Although our model shows a certain degree of robustness to moderate mask errors, extreme cases remain challenging. Moreover, the requirement for user-provided or precomputed masks may limit the method’s applicability in fully automatic pipelines or real-time settings.

We consider improving mask robustness and exploring mask-free alternatives as promising directions for future work.


%% file: aaai2026.bib
@article{long2025follow,
  title={Follow-your-shape: Shape-aware image editing via trajectory-guided region control},
  author={Long, Zeqian and Zheng, Mingzhe and Feng, Kunyu and Zhang, Xinhua and Liu, Hongyu and Yang, Harry and Zhang, Linfeng and Chen, Qifeng and Ma, Yue},
  journal={arXiv preprint arXiv:2508.08134},
  year={2025}
}

@inproceedings{zhang2025follow,
  title={Follow-Your-MultiPose: Tuning-Free Multi-Character Text-to-Video Generation via Pose Guidance},
  author={Zhang, Beiyuan and Ma, Yue and Fu, Chunlei and Song, Xinyang and Sun, Zhenan and Li, Ziqiang},
  booktitle={ICASSP 2025-2025 IEEE International Conference on Acoustics, Speech and Signal Processing (ICASSP)},
  pages={1--5},
  year={2025},
  organization={IEEE}
}

@article{zhao2023controlvideo,
  title={Controlvideo: Adding conditional control for one shot text-to-video editing},
  author={Zhao, Min and Wang, Rongzhen and Bao, Fan and Li, Chongxuan and Zhu, Jun},
  journal={arXiv preprint arXiv:2305.17098},
  volume={2},
  number={3},
  year={2023}
}

@inproceedings{wangtaming,
  title={Taming Rectified Flow for Inversion and Editing},
  author={Wang, Jiangshan and Pu, Junfu and Qi, Zhongang and Guo, Jiayi and Ma, Yue and Huang, Nisha and Chen, Yuxin and Li, Xiu and Shan, Ying},
  booktitle={Forty-second International Conference on Machine Learning},
year={2024}

}

@article{kong2024hunyuanvideo,
  title={Hunyuanvideo: A systematic framework for large video generative models},
  author={Kong, Weijie and Tian, Qi and Zhang, Zijian and Min, Rox and Dai, Zuozhuo and Zhou, Jin and Xiong, Jiangfeng and Li, Xin and Wu, Bo and Zhang, Jianwei and others},
  journal={arXiv preprint arXiv:2412.03603},
  year={2024}
}

@article{xing2024make,
  title={Make-your-video: Customized video generation using textual and structural guidance},
  author={Xing, Jinbo and Xia, Menghan and Liu, Yuxin and Zhang, Yuechen and Zhang, Yong and He, Yingqing and Liu, Hanyuan and Chen, Haoxin and Cun, Xiaodong and Wang, Xintao and others},
  journal={IEEE Transactions on Visualization and Computer Graphics},
  volume={31},
  number={2},
  pages={1526--1541},
  year={2024},
  publisher={IEEE}
}

@inproceedings{xing2024dynamicrafter,
  title={Dynamicrafter: Animating open-domain images with video diffusion priors},
  author={Xing, Jinbo and Xia, Menghan and Zhang, Yong and Chen, Haoxin and Yu, Wangbo and Liu, Hanyuan and Liu, Gongye and Wang, Xintao and Shan, Ying and Wong, Tien-Tsin},
  booktitle={European Conference on Computer Vision},
  pages={399--417},
  year={2024},
  organization={Springer}
}

@article{hu2024motionmaster,
  title={MotionMaster: Training-free Camera Motion Transfer For Video Generation},
  author={Hu, Teng and Zhang, Jiangning and Yi, Ran and Wang, Yating and Huang, Hongrui and Weng, Jieyu and Wang, Yabiao and Ma, Lizhuang},
  journal={CoRR},
  year={2024}
}

@inproceedings{pondaven2025video,
  title={Video motion transfer with diffusion transformers},
  author={Pondaven, Alexander and Siarohin, Aliaksandr and Tulyakov, Sergey and Torr, Philip and Pizzati, Fabio},
  booktitle={Proceedings of the Computer Vision and Pattern Recognition Conference},
  pages={22911--22921},
  year={2025}
}

@article{yesiltepe2024motionshop,
  title={MotionShop: Zero-Shot Motion Transfer in Video Diffusion Models with Mixture of Score Guidance},
  author={Yesiltepe, Hidir and Meral, Tuna Han Salih and Dunlop, Connor and Yanardag, Pinar},
  journal={CoRR},
  year={2024}
}

@inproceedings{zhao2024motiondirector,
  title={Motiondirector: Motion customization of text-to-video diffusion models},
  author={Zhao, Rui and Gu, Yuchao and Wu, Jay Zhangjie and Zhang, David Junhao and Liu, Jia-Wei and Wu, Weijia and Keppo, Jussi and Shou, Mike Zheng},
  booktitle={European Conference on Computer Vision},
  pages={273--290},
  year={2024},
  organization={Springer}
}

@inproceedings{songdenoising,
  title={Denoising Diffusion Implicit Models},
  author={Song, Jiaming and Meng, Chenlin and Ermon, Stefano},
  booktitle={International Conference on Learning Representations},
  year={2021}
}

@inproceedings{songscore,
  title={Score-Based Generative Modeling through Stochastic Differential Equations},
  author={Song, Yang and Sohl-Dickstein, Jascha and Kingma, Diederik P and Kumar, Abhishek and Ermon, Stefano and Poole, Ben},
  booktitle={International Conference on Learning Representations},
year={2020}
}

@article{lu2025dpm,
  title={Dpm-solver++: Fast solver for guided sampling of diffusion probabilistic models},
  author={Lu, Cheng and Zhou, Yuhao and Bao, Fan and Chen, Jianfei and Li, Chongxuan and Zhu, Jun},
  journal={Machine Intelligence Research},
  pages={1--22},
  year={2025},
  publisher={Springer}
}

@inproceedings{wallace2023edict,
  title={Edict: Exact diffusion inversion via coupled transformations},
  author={Wallace, Bram and Gokul, Akash and Naik, Nikhil},
  booktitle={Proceedings of the IEEE/CVF Conference on Computer Vision and Pattern Recognition},
  pages={22532--22541},
  year={2023}
}

@inproceedings{zhaounipc,
  title={UniPC: A Unified Predictor-Corrector Framework for Fast Sampling of Diffusion Models},
  author={Zhao, Wenliang and Bai, Lujia and Rao, Yongming and Zhou, Jie and Lu, Jiwen},
  booktitle={Thirty-seventh Conference on Neural Information Processing Systems},
year={2023}
}

@inproceedings{yatim2024space,
  title={Space-time diffusion features for zero-shot text-driven motion transfer},
  author={Yatim, Danah and Fridman, Rafail and Bar-Tal, Omer and Kasten, Yoni and Dekel, Tali},
  booktitle={Proceedings of the IEEE/CVF Conference on Computer Vision and Pattern Recognition},
  pages={8466--8476},
  year={2024}
}

@inproceedings{peebles2023scalable,
  title={Scalable diffusion models with transformers},
  author={Peebles, William and Xie, Saining},
  booktitle={Proceedings of the IEEE/CVF international conference on computer vision},
  pages={4195--4205},
  year={2023}
}

@article{xiao2024video,
  title={Video diffusion models are training-free motion interpreter and controller},
  author={Xiao, Zeqi and Zhou, Yifan and Yang, Shuai and Pan, Xingang},
  journal={Advances in Neural Information Processing Systems},
  volume={37},
  pages={76115--76138},
  year={2024}
}

@inproceedings{lingmotionclone,
  title={MotionClone: Training-Free Motion Cloning for Controllable Video Generation},
  author={Ling, Pengyang and Bu, Jiazi and Zhang, Pan and Dong, Xiaoyi and Zang, Yuhang and Wu, Tong and Chen, Huaian and Wang, Jiaqi and Jin, Yi},
  booktitle={The Thirteenth International Conference on Learning Representations},
year={2024}
}

@inproceedings{ravisam,
  title={SAM 2: Segment Anything in Images and Videos},
  author={Ravi, Nikhila and Gabeur, Valentin and Hu, Yuan-Ting and Hu, Ronghang and Ryali, Chaitanya and Ma, Tengyu and Khedr, Haitham and R{\"a}dle, Roman and Rolland, Chloe and Gustafson, Laura and others},
  booktitle={The Thirteenth International Conference on Learning Representations}
}

@article{ma2025controllable,
  title={Controllable Video Generation: A Survey},
  author={Ma, Yue and Feng, Kunyu and Hu, Zhongyuan and Wang, Xinyu and Wang, Yucheng and Zheng, Mingzhe and He, Xuanhua and Zhu, Chenyang and Liu, Hongyu and He, Yingqing and others},
  journal={arXiv preprint arXiv:2507.16869},
  year={2025}
}

@article{ma2025followyourmotion,
  title={Follow-Your-Motion: Video Motion Transfer via Efficient Spatial-Temporal Decoupled Finetuning},
  author={Ma, Yue and Liu, Yulong and Zhu, Qiyuan and Yang, Ayden and Feng, Kunyu and Zhang, Xinhua and Li, Zhifeng and Han, Sirui and Qi, Chenyang and Chen, Qifeng},
  journal={arXiv preprint arXiv:2506.05207},
  year={2025}
}

@article{zhang2025magiccolor,
  title={MagicColor: Multi-instance sketch colorization},
  author={Zhang, Yinhan and Ma, Yue and Wang, Bingyuan and Chen, Qifeng and Wang, Zeyu},
  journal={arXiv preprint arXiv:2503.16948},
  year={2025}
}

@article{ma2025followfaster,
  title={Follow-your-emoji-faster: Towards efficient, fine-controllable, and expressive freestyle portrait animation},
  author={Ma, Yue and Yan, Zexuan and Liu, Hongyu and Wang, Hongfa and Pan, Heng and He, Yingqing and Yuan, Junkun and Zeng, Ailing and Cai, Chengfei and Shum, Heung-Yeung and others},
  journal={arXiv preprint arXiv:2509.16630},
  year={2025}
}

@inproceedings{ma2024followyouremoji,
  title={Follow-your-emoji: Fine-controllable and expressive freestyle portrait animation},
  author={Ma, Yue and Liu, Hongyu and Wang, Hongfa and Pan, Heng and He, Yingqing and Yuan, Junkun and Zeng, Ailing and Cai, Chengfei and Shum, Heung-Yeung and Liu, Wei and others},
  booktitle={SIGGRAPH Asia 2024 Conference Papers},
  pages={1--12},
  year={2024}
}

@article{ma2023magicstick,
  title={Magicstick: Controllable video editing via control handle transformations},
  author={Ma, Yue and Cun, Xiaodong and He, Yingqing and Qi, Chenyang and Wang, Xintao and Shan, Ying and Li, Xiu and Chen, Qifeng},
  journal={arXiv preprint arXiv:2312.03047},
  year={2023}
}

@article{yuluo2025gr,
  title={GR-Gaussian: Graph-Based Radiative Gaussian Splatting for Sparse-View CT Reconstruction},
  author={Yuluo, Yikuang and Ma, Yue and Shen, Kuan and Jin, Tongtong and Liao, Wang and Ma, Yangpu and Wang, Fuquan},
  journal={arXiv preprint arXiv:2508.02408},
  year={2025}
}

@article{chen2025contextflow,
  title={ContextFlow: Training-Free Video Object Editing via Adaptive Context Enrichment},
  author={Chen, Yiyang and He, Xuanhua and Ma, Xiujun and Ma, Yue},
  journal={arXiv preprint arXiv:2509.17818},
  year={2025}
}

@inproceedings{song2025idprotector,
  title={Idprotector: An adversarial noise encoder to protect against id-preserving image generation},
  author={Song, Yiren and Yang, Pei and Ci, Hai and Shou, Mike Zheng},
  booktitle={Proceedings of the Computer Vision and Pattern Recognition Conference},
  pages={3019--3028},
  year={2025}
}

@article{song2024anti,
  title={Anti-Reference: Universal and Immediate Defense Against Reference-Based Generation},
  author={Song, Yiren and Lou, Shengtao and Liu, Xiaokang and Ci, Hai and Yang, Pei and Liu, Jiaming and Shou, Mike Zheng},
  journal={arXiv preprint arXiv:2412.05980},
  year={2024}
}

@article{ci2024wmadapter,
  title={Wmadapter: Adding watermark control to latent diffusion models},
  author={Ci, Hai and Song, Yiren and Yang, Pei and Xie, Jinheng and Shou, Mike Zheng},
  journal={arXiv preprint arXiv:2406.08337},
  year={2024}
}

@article{guo2023animatediff,
  title={AnimateDiff: Animate Your Personalized Text-to-Image Diffusion Models without Specific Tuning},
  author={Guo, Yuwei and Yang, Ceyuan and Rao, Anyi and Liang, Zhengyang and Wang, Yaohui and Qiao, Yu and Agrawala, Maneesh and Lin, Dahua and Dai, Bo},
  journal={International Conference on Learning Representations},
  year={2024}
}

@inproceedings{ma2022visual,
  title={Visual knowledge graph for human action reasoning in videos},
  author={Ma, Yue and Wang, Yali and Wu, Yue and Lyu, Ziyu and Chen, Siran and Li, Xiu and Qiao, Yu},
  booktitle={Proceedings of the 30th ACM International Conference on Multimedia},
  pages={4132--4141},
  year={2022}
}

@inproceedings{ma2025followyourclick,
  title={Follow-your-click: Open-domain regional image animation via motion prompts},
  author={Ma, Yue and He, Yingqing and Wang, Hongfa and Wang, Andong and Shen, Leqi and Qi, Chenyang and Ying, Jixuan and Cai, Chengfei and Li, Zhifeng and Shum, Heung-Yeung and others},
  booktitle={Proceedings of the AAAI Conference on Artificial Intelligence},
  volume={39},
  number={6},
  pages={6018--6026},
  year={2025}
}

@inproceedings{liu2025avatarartist,
  title={Avatarartist: Open-domain 4d avatarization},
  author={Liu, Hongyu and Wang, Xuan and Wan, Ziyu and Ma, Yue and Chen, Jingye and Fan, Yanbo and Shen, Yujun and Song, Yibing and Chen, Qifeng},
  booktitle={Proceedings of the Computer Vision and Pattern Recognition Conference},
  pages={10758--10769},
  year={2025}
}

@inproceedings{yang2023uni,
  title={Uni-paint: A unified framework for multimodal image inpainting with pretrained diffusion model},
  author={Yang, Shiyuan and Chen, Xiaodong and Liao, Jing},
  booktitle={Proceedings of the 31st ACM International Conference on Multimedia},
  pages={3190--3199},
  year={2023}
}

@inproceedings{yangcogvideox,
  title={CogVideoX: Text-to-Video Diffusion Models with An Expert Transformer},
  author={Yang, Zhuoyi and Teng, Jiayan and Zheng, Wendi and Ding, Ming and Huang, Shiyu and Xu, Jiazheng and Yang, Yuanming and Hong, Wenyi and Zhang, Xiaohan and Feng, Guanyu and others},
  booktitle={The Thirteenth International Conference on Learning Representations},
year={2024}
}

@article{xu2024easyanimate,
  title={Easyanimate: A high-performance long video generation method based on transformer architecture},
  author={Xu, Jiaqi and Zou, Xinyi and Huang, Kunzhe and Chen, Yunkuo and Liu, Bo and Cheng, MengLi and Shi, Xing and Huang, Jun},
  journal={arXiv preprint arXiv:2405.18991},
  year={2024}
}

@article{wan2025wan,
  title={Wan: Open and advanced large-scale video generative models},
  author={Wan, Team and Wang, Ang and Ai, Baole and Wen, Bin and Mao, Chaojie and Xie, Chen-Wei and Chen, Di and Yu, Feiwu and Zhao, Haiming and Yang, Jianxiao and others},
  journal={arXiv preprint arXiv:2503.20314},
  year={2025}
}

@article{liu2024sora,
  title={Sora: A Review on Background, Technology, Limitations, and Opportunities of Large Vision Models},
  author={Liu, Yixin and Zhang, Kai and Li, Yuan and Yan, Zhiling and Gao, Chujie and Chen, Ruoxi and Yuan, Zhengqing and Huang, Yue and Sun, Hanchi and Gao, Jianfeng and others},
  journal={CoRR},
  year={2024}
}

@article{zhang2025framepainter,
  title={Framepainter: Endowing interactive image editing with video diffusion priors},
  author={Zhang, Yabo and Zhou, Xinpeng and Zeng, Yihan and Xu, Hang and Li, Hui and Zuo, Wangmeng},
  journal={arXiv preprint arXiv:2501.08225},
  year={2025}
}

@inproceedings{xu2025hunyuanportrait,
  title={Hunyuanportrait: Implicit condition control for enhanced portrait animation},
  author={Xu, Zunnan and Yu, Zhentao and Zhou, Zixiang and Zhou, Jun and Jin, Xiaoyu and Hong, Fa-Ting and Ji, Xiaozhong and Zhu, Junwei and Cai, Chengfei and Tang, Shiyu and others},
  booktitle={Proceedings of the Computer Vision and Pattern Recognition Conference},
  pages={15909--15919},
  year={2025}
}

@inproceedings{jeong2024dreammotion,
  title={Dreammotion: Space-time self-similar score distillation for zero-shot video editing},
  author={Jeong, Hyeonho and Chang, Jinho and Park, Geon Yeong and Ye, Jong Chul},
  booktitle={European Conference on Computer Vision},
  pages={358--376},
  year={2024},
  organization={Springer}
}

@article{elarabawy2022direct,
  title={Direct inversion: Optimization-free text-driven real image editing with diffusion models},
  author={Elarabawy, Adham and Kamath, Harish and Denton, Samuel},
  journal={arXiv preprint arXiv:2211.07825},
  year={2022}
}

@inproceedings{miyake2025negative,
  title={Negative-prompt inversion: Fast image inversion for editing with text-guided diffusion models},
  author={Miyake, Daiki and Iohara, Akihiro and Saito, Yu and Tanaka, Toshiyuki},
  booktitle={2025 IEEE/CVF Winter Conference on Applications of Computer Vision (WACV)},
  pages={2063--2072},
  year={2025},
  organization={IEEE}
}

@inproceedings{rout2024beyond,
  title={Beyond First-Order Tweedie: Solving Inverse Problems using Latent Diffusion},
  author={Rout, Litu and Chen, Yujia and Kumar, Abhishek and Caramanis, Constantine and Shakkottai, Sanjay and Chu, Wen-Sheng},
  booktitle={2024 IEEE/CVF Conference on Computer Vision and Pattern Recognition (CVPR)},
  pages={9472--9481},
  year={2024},
  organization={IEEE}
}

@inproceedings{routsemantic,
  title={Semantic Image Inversion and Editing using Rectified Stochastic Differential Equations},
  author={Rout, Litu and Chen, Yujia and Ruiz, Nataniel and Caramanis, Constantine and Shakkottai, Sanjay and Chu, Wen-Sheng},
  booktitle={The Thirteenth International Conference on Learning Representations},
  year={2024}
}

@inproceedings{kingma2014adam,
  title={Adam: a method for stochastic optimization},
  author={Kingma, DP},
  booktitle={Int Conf Learn Represent},
  year={2014}
}

@inproceedings{wang2025motion,
  title={Motion inversion for video customization},
  author={Wang, Luozhou and Mai, Ziyang and Shen, Guibao and Liang, Yixun and Tao, Xin and Wan, Pengfei and Zhang, Di and Li, Yijun and Chen, Ying-Cong},
  booktitle={Proceedings of the Special Interest Group on Computer Graphics and Interactive Techniques Conference Conference Papers},
  pages={1--12},
  year={2025}
}

@article{zhang2023motioncrafter,
  title={Motioncrafter: One-shot motion customization of diffusion models},
  author={Zhang, Yuxin and Tang, Fan and Huang, Nisha and Huang, Haibin and Ma, Chongyang and Dong, Weiming and Xu, Changsheng},
  journal={arXiv preprint arXiv:2312.05288},
  year={2023}
}

@article{zhang2025easycontrol,
  title={Easycontrol: Adding efficient and flexible control for diffusion transformer},
  author={Zhang, Yuxuan and Yuan, Yirui and Song, Yiren and Wang, Haofan and Liu, Jiaming},
  journal={arXiv preprint arXiv:2503.07027},
  year={2025}
}

@inproceedings{zhang2024ssr,
  title={Ssr-encoder: Encoding selective subject representation for subject-driven generation},
  author={Zhang, Yuxuan and Song, Yiren and Liu, Jiaming and Wang, Rui and Yu, Jinpeng and Tang, Hao and Li, Huaxia and Tang, Xu and Hu, Yao and Pan, Han and others},
  booktitle={Proceedings of the IEEE/CVF Conference on Computer Vision and Pattern Recognition},
  pages={8069--8078},
  year={2024}
}

@article{song2025layertracer,
  title={LayerTracer: Cognitive-Aligned Layered SVG Synthesis via Diffusion Transformer},
  author={Song, Yiren and Chen, Danze and Shou, Mike Zheng},
  journal={arXiv preprint arXiv:2502.01105},
  year={2025}
}

@article{song2025makeanything,
  title={MakeAnything: Harnessing Diffusion Transformers for Multi-Domain Procedural Sequence Generation},
  author={Song, Yiren and Liu, Cheng and Shou, Mike Zheng},
  journal={arXiv preprint arXiv:2502.01572},
  year={2025}
}

@article{huang2025photodoodle,
  title={Photodoodle: Learning artistic image editing from few-shot pairwise data},
  author={Huang, Shijie and Song, Yiren and Zhang, Yuxuan and Guo, Hailong and Wang, Xueyin and Shou, Mike Zheng and Liu, Jiaming},
  journal={arXiv preprint arXiv:2502.14397},
  year={2025}
}

@String(CVPR= {IEEE Conf. Comput. Vis. Pattern Recog.})

@String(ICCV= {Int. Conf. Comput. Vis.})

@String(ICASSP=	{ICASSP})

@String(AAAI = {AAAI})

@String(CVPR  = {CVPR})

@String(ICCV  = {ICCV})

@inproceedings{rombach2022high,
  title={High-resolution image synthesis with latent diffusion models},
  author={Rombach, Robin and Blattmann, Andreas and Lorenz, Dominik and Esser, Patrick and Ommer, Bj{\"o}rn},
  booktitle={Proceedings of the IEEE/CVF Conference on Computer Vision and Pattern Recognition},
  pages={10684--10695},
  year={2022}
}

@article{mokady2023null,
  title={Null-text inversion for precise text-to-image editing},
  author={Mokady, Ron and Hertz, Amir and Aberman, Kfir and Pritch, Yael and Cohen-Or, Daniel},
  journal={arXiv preprint arXiv:2211.09794},
  year={2023}
}

@article{ma2025followcreation,
  title={Follow-Your-Creation: Empowering 4D Creation through Video Inpainting},
  author={Ma, Yue and Feng, Kunyu and Zhang, Xinhua and Liu, Hongyu and Zhang, David Junhao and Xing, Jinbo and Zhang, Yinhan and Yang, Ayden and Wang, Zeyu and Chen, Qifeng},
  journal={arXiv preprint arXiv:2506.04590},
  year={2025}
}

@article{yan2025eedit,
  title={Eedit: Rethinking the spatial and temporal redundancy for efficient image editing},
  author={Yan, Zexuan and Ma, Yue and Zou, Chang and Chen, Wenteng and Chen, Qifeng and Zhang, Linfeng},
  journal={arXiv preprint arXiv:2503.10270},
  year={2025}
}

@article{zhu2024instantswap,
  title={Instantswap: Fast customized concept swapping across sharp shape differences},
  author={Zhu, Chenyang and Li, Kai and Ma, Yue and Tang, Longxiang and Fang, Chengyu and Chen, Chubin and Chen, Qifeng and Li, Xiu},
  journal={arXiv preprint arXiv:2412.01197},
  year={2024}
}

@article{wang2024cove,
  title={Cove: Unleashing the diffusion feature correspondence for consistent video editing},
  author={Wang, Jiangshan and Ma, Yue and Guo, Jiayi and Xiao, Yicheng and Huang, Gao and Li, Xiu},
  journal={arXiv preprint arXiv:2406.08850},
  year={2024}
}

@article{chen2024follow,
  title={Follow-your-canvas: Higher-resolution video outpainting with extensive content generation},
  author={Chen, Qihua and Ma, Yue and Wang, Hongfa and Yuan, Junkun and Zhao, Wenzhe and Tian, Qi and Wang, Hongmei and Min, Shaobo and Chen, Qifeng and Liu, Wei},
  journal={arXiv preprint arXiv:2409.01055},
  year={2024}
}

@article{feng2025follow,
  title={Follow-your-instruction: A comprehensive mllm agent for world data synthesis},
  author={Feng, Kunyu and Ma, Yue and Zhang, Xinhua and Liu, Boshi and Yuluo, Yikuang and Zhang, Yinhan and Liu, Runtao and Liu, Hongyu and Qin, Zhiyuan and Mo, Shanhui and others},
  journal={arXiv preprint arXiv:2508.05580},
  year={2025}
}

@inproceedings{zhu2025multibooth,
  title={Multibooth: Towards generating all your concepts in an image from text},
  author={Zhu, Chenyang and Li, Kai and Ma, Yue and He, Chunming and Li, Xiu},
  booktitle={Proceedings of the AAAI Conference on Artificial Intelligence},
  volume={39},
  number={10},
  pages={10923--10931},
  year={2025}
}

@inproceedings{feng2025dit4edit,
  title={Dit4edit: Diffusion transformer for image editing},
  author={Feng, Kunyu and Ma, Yue and Wang, Bingyuan and Qi, Chenyang and Chen, Haozhe and Chen, Qifeng and Wang, Zeyu},
  booktitle={Proceedings of the AAAI Conference on Artificial Intelligence},
  volume={39},
  number={3},
  pages={2969--2977},
  year={2025}
}

@inproceedings{ma2024followpose,
  title={Follow your pose: Pose-guided text-to-video generation using pose-free videos},
  author={Ma, Yue and He, Yingqing and Cun, Xiaodong and Wang, Xintao and Chen, Siran and Li, Xiu and Chen, Qifeng},
  booktitle={Proceedings of the AAAI Conference on Artificial Intelligence},
  volume={38},
  number={5},
  pages={4117--4125},
  year={2024}
}

@inproceedings{sun2025gl,
  title={GL-LCM: Global-Local Latent Consistency Models for Fast High-Resolution Bone Suppression in Chest X-Ray Images},
  author={Sun, Yifei and Chen, Zhanghao and Zheng, Hao and Lu, Yuqing and Duan, Lixin and Fan, Fenglei and Elazab, Ahmed and Wan, Xiang and Wang, Changmiao and Ge, Ruiquan},
  booktitle={International Conference on Medical Image Computing and Computer-Assisted Intervention},
  pages={222--232},
  year={2025},
  organization={Springer}
}

@article{sun2025bs,
  title={BS-LDM: Effective Bone Suppression in High-Resolution Chest X-Ray Images with Conditional Latent Diffusion Models},
  author={Sun, Yifei and Chen, Zhanghao and Zheng, Hao and Deng, Wenming and Liu, Jin and Min, Wenwen and Elazab, Ahmed and Wan, Xiang and Wang, Changmiao and Ge, Ruiquan},
  journal={IEEE Journal of Biomedical and Health Informatics},
  year={2025},
  publisher={IEEE}
}

@inproceedings{zhang2025echomask,
  title={EchoMask: Speech-Queried Attention-based Mask Modeling for Holistic Co-Speech Motion Generation},
  author={Zhang, Xiangyue and Li, Jianfang and Zhang, Jiaxu and Ren, Jianqiang and Bo, Liefeng and Tu, Zhigang},
  booktitle={Proceedings of the 33rd ACM International Conference on Multimedia},
  pages={10827--10836},
  year={2025}
}

@inproceedings{zhang2025semtalk,
  title={SemTalk: Holistic Co-speech Motion Generation with Frame-level Semantic Emphasis},
  author={Zhang, Xiangyue and Li, Jianfang and Zhang, Jiaxu and Dang, Ziqiang and Ren, Jianqiang and Bo, Liefeng and Tu, Zhigang},
  booktitle={Proceedings of the IEEE/CVF International Conference on Computer Vision},
  pages={13761--13771},
  year={2025}
}

@article{wang2025characterfactory,
  title={Characterfactory: Sampling consistent characters with gans for diffusion models},
  author={Wang, Qinghe and Li, Baolu and Li, Xiaomin and Cao, Bing and Ma, Liqian and Lu, Huchuan and Jia, Xu},
  journal={IEEE Transactions on Image Processing},
  year={2025},
  publisher={IEEE}
}

@InProceedings{zhao2025unified,
    title={A Unified Solution to Video Fusion: From Multi-Frame Learning to Benchmarking},
    author={Zhao, Zixiang and Bai, Haowen and Ke, Bingxin and Cui, Yukun and Deng, Lilun and Zhang, Yulun and Zhang, Kai and Schindler, Konrad},
    booktitle = {Advances in Neural Information Processing Systems (NeurIPS)},
    year={2025}
}

@InProceedings{Zhao_2023_CVPR,
    author    = {Zhao, Zixiang and Bai, Haowen and Zhang, Jiangshe and Zhang, Yulun and Xu, Shuang and Lin, Zudi and Timofte, Radu and Van Gool, Luc},
    title     = {CDDFuse: Correlation-Driven Dual-Branch Feature Decomposition for Multi-Modality Image Fusion},
    booktitle = {Proceedings of the IEEE/CVF Conference on Computer Vision and Pattern Recognition (CVPR)},
    month     = {June},
    year      = {2023},
    pages     = {5906-5916}
}

@InProceedings{Zhao_2023_ICCV_DDFM,
    author    = {Zhao, Zixiang and Bai, Haowen and Zhu, Yuanzhi and Zhang, Jiangshe and Xu, Shuang and Zhang, Yulun and Zhang, Kai and Meng, Deyu and Timofte, Radu and Van Gool, Luc},
    title     = {DDFM: Denoising Diffusion Model for Multi-Modality Image Fusion},
    booktitle = {Proceedings of the IEEE/CVF International Conference on Computer Vision (ICCV)},
    month     = {October},
    year      = {2023},
    pages     = {8082-8093}
}

@inproceedings{liu2023delving,  
title={Delving stylegan inversion for image editing: A foundation latent space viewpoint},  
author={Liu, Hongyu and Song, Yibing and Chen, Qifeng},  
booktitle={Proceedings of the IEEE/CVF conference on computer vision and pattern recognition},  
pages={10072--10082},  
year={2023}
}

@inproceedings{liu2024headartist,  
title={Headartist: Text-conditioned 3d head generation with self score distillation},  
author={Liu, Hongyu and Wang, Xuan and Wan, Ziyu and Shen, Yujun and Song, Yibing and Liao, Jing and Chen, Qifeng},  
booktitle={ACM SIGGRAPH 2024 Conference Papers},  pages={1--12},  
year={2024}}

@article{liu2023human,  title={Human motionformer: Transferring human motions with vision transformers},  author={Liu, Hongyu and Han, Xintong and Jin, Chengbin and Qian, Lihui and Wei, Huawei and Lin, Zhe and Wang, Faqiang and Dong, Haoye and Song, Yibing and Xu, Jia and others},  journal={arXiv preprint arXiv:2302.11306},  
year={2023}
}

@article{zhu2022one,
title={One model to edit them all: Free-form text-driven image manipulation with semantic modulations},  
author={Zhu, Yiming and Liu, Hongyu and Song, Yibing and Yuan, Ziyang and Han, Xintong and Yuan, Chun and Chen, Qifeng and Wang, Jue},  journal={Advances in Neural Information Processing Systems},  
volume={35},  
pages={25146--25159},  
year={2022}
}

@inproceedings{feng20254dgs,
  title={E-4DGS: High-Fidelity Dynamic Reconstruction from the Multi-view Event Cameras},
  author={Feng, Chaoran and Tang, Zhenyu and Yu, Wangbo and Pang, Yatian and Zhao, Yian and Zhao, Jianbin and Yuan, Li and Tian, Yonghong},
  booktitle={Proceedings of the 33rd ACM International Conference on Multimedia},
  pages={7356--7365},
  year={2025}
}

@inproceedings{li2025gs2e,
  title={GS2E: Gaussian Splatting is an Effective Data Generator for Event Stream Generation},
  author={Li*, Yuchen and Feng*, Chaoran and Tang, Zhenyu and Deng, Kaiyuan and Yu, Wangbo and Tian, Yonghong and Yuan, Li},
  booktitle={NeurIPS 2025},
  year={2025}
}

@inproceedings{jia2019comdefend,
  title={Comdefend: An efficient image compression model to defend adversarial examples},
  author={Jia, Xiaojun and Wei, Xingxing and Cao, Xiaochun and Foroosh, Hassan},
  booktitle={Proceedings of the IEEE/CVF conference on computer vision and pattern recognition},
  pages={6084--6092},
  year={2019}
}

@inproceedings{jia2022adversarial,
  title={LAS-AT: adversarial training with learnable attack strategy},
  author={Jia, Xiaojun and Zhang, Yong and Wu, Baoyuan and Ma, Ke and Wang, Jue and Cao, Xiaochun},
  booktitle={Proceedings of the IEEE/CVF Conference on Computer Vision and Pattern Recognition},
  pages={13398--13408},
  year={2022}
}

@article{jia2022boosting,
  title={Boosting fast adversarial training with learnable adversarial initialization},
  author={Jia, Xiaojun and Zhang, Yong and Wu, Baoyuan and Wang, Jue and Cao, Xiaochun},
  journal={IEEE Transactions on Image Processing},
  volume={31},
  pages={4417--4430},
  year={2022},
  publisher={IEEE}
}

@inproceedings{
Z1,
title={{KABB}: Knowledge-Aware Bayesian Bandits for Dynamic Expert Coordination in Multi-Agent Systems},
author={Jusheng Zhang and Zimeng Huang and Yijia Fan and Ningyuan Liu and Mingyan Li and Zhuojie Yang and Jiawei Yao and Jian Wang and Keze Wang},
booktitle={Forty-second International Conference on Machine Learning},
year={2025},
url={https://openreview.net/forum?id=AKvy9a4jho}
}

@misc{Z2,
      title={GAM-Agent: Game-Theoretic and Uncertainty-Aware Collaboration for Complex Visual Reasoning}, 
      author={Jusheng Zhang and Yijia Fan and Wenjun Lin and Ruiqi Chen and Haoyi Jiang and Wenhao Chai and Jian Wang and Keze Wang},
      year={2025},
      eprint={2505.23399},
      archivePrefix={arXiv},
      primaryClass={cs.AI},
      url={https://arxiv.org/abs/2505.23399}, 
}

@misc{Z3,
      title={CF-VLM:CounterFactual Vision-Language Fine-tuning}, 
      author={Jusheng Zhang and Kaitong Cai and Yijia Fan and Jian Wang and Keze Wang},
      year={2025},
      eprint={2506.17267},
      archivePrefix={arXiv},
      primaryClass={cs.LG},
      url={https://arxiv.org/abs/2506.17267}, 
}

@misc{Z4,
      title={Kolmogorov-Arnold Fourier Networks}, 
      author={Jusheng Zhang and Yijia Fan and Kaitong Cai and Keze Wang},
      year={2025},
      eprint={2502.06018},
      archivePrefix={arXiv},
      primaryClass={cs.LG},
      url={https://arxiv.org/abs/2502.06018}, 
}

@misc{Z5,
      title={OSC: Cognitive Orchestration through Dynamic Knowledge Alignment in Multi-Agent LLM Collaboration}, 
      author={Jusheng Zhang and Yijia Fan and Kaitong Cai and Xiaofei Sun and Keze Wang},
      year={2025},
      eprint={2509.04876},
      archivePrefix={arXiv},
      primaryClass={cs.AI},
      url={https://arxiv.org/abs/2509.04876}, 
}

@misc{MAT,
      title={MAT-Agent: Adaptive Multi-Agent Training Optimization}, 
      author={Jusheng Zhang and Kaitong Cai and Yijia Fan and Ningyuan Liu and Keze Wang},
      year={2025},
      eprint={2510.17845},
      archivePrefix={arXiv},
      primaryClass={cs.CV},
      url={https://arxiv.org/abs/2510.17845}, 
}

@misc{Z6,
      title={DrDiff: Dynamic Routing Diffusion with Hierarchical Attention for Breaking the Efficiency-Quality Trade-off}, 
      author={Jusheng Zhang and Yijia Fan and Kaitong Cai and Zimeng Huang and Xiaofei Sun and Jian Wang and Chengpei Tang and Keze Wang},
      year={2025},
      eprint={2509.02785},
      archivePrefix={arXiv},
      primaryClass={cs.CL},
      url={https://arxiv.org/abs/2509.02785}, 
}

@misc{Z7,
      title={Learning Dynamics of VLM Finetuning}, 
      author={Jusheng Zhang and Kaitong Cai and Jing Yang and Keze Wang},
      year={2025},
      eprint={2510.11978},
      archivePrefix={arXiv},
      primaryClass={cs.LG},
      url={https://arxiv.org/abs/2510.11978}, 
}

@misc{Z8,
      title={Failure-Driven Workflow Refinement}, 
      author={Jusheng Zhang and Kaitong Cai and Qinglin Zeng and Ningyuan Liu and Stephen Fan and Ziliang Chen and Keze Wang},
      year={2025},
      eprint={2510.10035},
      archivePrefix={arXiv},
      primaryClass={cs.AI},
      url={https://arxiv.org/abs/2510.10035}, 
}

@misc{Z9,
      title={Top-Down Semantic Refinement for Image Captioning}, 
      author={Jusheng Zhang and Kaitong Cai and Jing Yang and Jian Wang and Chengpei Tang and Keze Wang},
      year={2025},
      eprint={2510.22391},
      archivePrefix={arXiv},
      primaryClass={cs.CV},
      url={https://arxiv.org/abs/2510.22391}, 
}

@article{she2025customvideox,
  title={CustomVideoX: 3D Reference Attention Driven Dynamic Adaptation for Zero-Shot Customized Video Diffusion Transformers},
  author={She, D and Liu, Mushui and Pang, Jingxuan and Wang, Jin and Yang, Zhen and He, Wanggui and Zhang, Guanghao and Wang, Yi and Huang, Qihan and Tang, Haobin and others},
  journal={arXiv preprint arXiv:2502.06527},
  year={2025}
}

@article{wang2025mint,
  title={Mint: Multi-modal chain of thought in unified generative models for enhanced image generation},
  author={Wang, Yi and Liu, Mushui and He, Wanggui and Zhang, Longxiang and Huang, Ziwei and Zhang, Guanghao and Shu, Fangxun and Tao, Zhong and She, Dong and Yu, Zhelun and others},
  journal={arXiv preprint arXiv:2503.01298},
  year={2025}
}

@article{ying2024restorerid,
  title={RestorerID: Towards Tuning-Free Face Restoration with ID Preservation},
  author={Ying, Jiacheng and Liu, Mushui and Wu, Zhe and Zhang, Runming and Yu, Zhu and Fu, Siming and Cao, Si-Yuan and Wu, Chao and Yu, Yunlong and Shen, Hui-Liang},
  journal={arXiv preprint arXiv:2411.14125},
  year={2024}
}

@inproceedings{liu2024llm4gen,
  title={TFCustom: Customized Image Generation with Time-Aware Frequency Feature Guidance},
  author={Mushui Liu and Dong She and Jingxuan Pang and et al and Yuanlei Hou and Siming Fu},
booktitle={CVPR},
year={2025}
}

@article{shen2025follow,
  title={Follow-Your-Preference: Towards Preference-Aligned Image Inpainting},
  author={Shen, Yutao and Yuan, Junkun and Aonishi, Toru and Nakayama, Hideki and Ma, Yue},
  journal={arXiv preprint arXiv:2509.23082},
  year={2025}
}

@inproceedings{wan2025unipaint,   
title={Unipaint: Unified space-time video inpainting via mixture-of-experts},   
author={Wan, Zhen and Qi, Chenyang and Liu, Zhiheng and Gui, Tao and Ma, Yue},   
booktitle={Proceedings of the IEEE/CVF International Conference on Computer Vision},   pages={1861--1871},   
year={2025} }

@article{hui2025autoregressive,
  title={Autoregressive Images Watermarking through Lexical Biasing: An Approach Resistant to Regeneration Attack},
  author={Hui, Siqi and Song, Yiren and Zhou, Sanping and Deng, Ye and Huang, Wenli and Wang, Jinjun},
  journal={arXiv preprint arXiv:2506.01011},
  year={2025}
}

@article{zhang2025zero,
  title={Zero-shot 3D-Aware Trajectory-Guided image-to-video generation via Test-Time Training},
  author={Zhang, Ruicheng and Zhou, Jun and Xu, Zunnan and Liu, Zihao and Huang, Jiehui and Zhang, Mingyang and Sun, Yu and Li, Xiu},
  journal={arXiv preprint arXiv:2509.06723},
  year={2025}
}

@inproceedings{ci2024ringid,
  title={Ringid: Rethinking tree-ring watermarking for enhanced multi-key identification},
  author={Ci, Hai and Yang, Pei and Song, Yiren and Shou, Mike Zheng},
  booktitle={European Conference on Computer Vision},
  pages={338--354},
  year={2024},
  organization={Springer}
}

@article{liu2024image,
  title={Image watermarks are removable using controllable regeneration from clean noise},
  author={Liu, Yepeng and Song, Yiren and Ci, Hai and Zhang, Yu and Wang, Haofan and Shou, Mike Zheng and Bu, Yuheng},
  journal={arXiv preprint arXiv:2410.05470},
  year={2024}
}

@article{yang2024can,
  title={Can simple averaging defeat modern watermarks?},
  author={Yang, Pei and Ci, Hai and Song, Yiren and Shou, Mike Zheng},
  journal={Advances in Neural Information Processing Systems},
  volume={37},
  pages={56644--56673},
  year={2024}
}

@inproceedings{shenlong,
  title={Long-Term TalkingFace Generation via Motion-Prior Conditional Diffusion Model},
  author={Shen, Fei and Wang, Cong and Gao, Junyao and Guo, Qin and Dang, Jisheng and Tang, Jinhui and Chua, Tat-Seng},
  booktitle={Forty-second International Conference on Machine Learning}
}

@article{shen2024imagpose,
  title={Imagpose: A unified conditional framework for pose-guided person generation},
  author={Shen, Fei and Tang, Jinhui},
  journal={Advances in neural information processing systems},
  volume={37},
  pages={6246--6266},
  year={2024}
}

@inproceedings{shen2025imagdressing,
  title={Imagdressing-v1: Customizable virtual dressing},
  author={Shen, Fei and Jiang, Xin and He, Xin and Ye, Hu and Wang, Cong and Du, Xiaoyu and Li, Zechao and Tang, Jinhui},
  booktitle={Proceedings of the AAAI Conference on Artificial Intelligence},
  volume={39},
  number={7},
  pages={6795--6804},
  year={2025}
}

@inproceedings{shen2024advancing,
title={Advancing Pose-Guided Image Synthesis with Progressive Conditional Diffusion Models},
author={Fei Shen and Hu Ye and Jun Zhang and Cong Wang and Xiao Han and Yang Wei},
booktitle={The Twelfth International Conference on Learning Representations},
year={2024},
url={https://openreview.net/forum?id=rHzapPnCgT}
}

@article{shen2025imaggarment,
  title={IMAGGarment-1: Fine-Grained Garment Generation for Controllable Fashion Design},
  author={Shen, Fei and Yu, Jian and Wang, Cong and Jiang, Xin and Du, Xiaoyu and Tang, Jinhui},
  journal={arXiv preprint arXiv:2504.13176},
  year={2025}
}

@article{shen2025imagedit,
  title={IMAGEdit: Let Any Subject Transform},
  author={Shen, Fei and Xu, Weihao and Yan, Rui and Zhang, Dong and Shu, Xiangbo and Tang, Jinhui},
  journal={arXiv preprint arXiv:2510.01186},
  year={2025}
}

@article{shen2025imagharmony,
  title={IMAGHarmony: Controllable Image Editing with Consistent Object Quantity and Layout},
  author={Shen, Fei and Du, Xiaoyu and Gao, Yutong and Yu, Jian and Cao, Yushe and Lei, Xing and Tang, Jinhui},
  journal={arXiv preprint arXiv:2506.01949},
  year={2025}
}
